\documentclass[letterpaper, 10 pt, conference]{ieeeconf}
\IEEEoverridecommandlockouts  
\usepackage{xcolor}
\usepackage{graphicx}
\usepackage{subcaption}   
\usepackage[misc]{ifsym}
\usepackage{amsmath,amsfonts}
\usepackage{algorithmic}
\usepackage{array}
\usepackage[compatibility=false]{caption}

\usepackage{physics}
\usepackage{mathtools}
\usepackage{stfloats}
\usepackage{url}
\usepackage{verbatim}
\usepackage{booktabs}
\usepackage{graphicx}
\usepackage{hyperref}
\usepackage{multirow}
\usepackage{comment}
\usepackage{graphicx}
\usepackage{dcolumn}
\usepackage{bm}
\usepackage{ulem}
\usepackage{boldline}

\newcommand{\myparagraph}[1]{\vspace{2pt}\noindent{\bf #1}}
\usepackage{pifont}

\usepackage[dvipsnames]{xcolor}

\title{\LARGE \bf
Vision-Guided Targeted Grasping and Vibration for\\Robotic Pollination in Controlled Environments
}

\author{Jaehwan Jeong$^{1,2,*}$, Tuan-Anh Vu$^{2,*}$, Radha Lahoti$^{2}$, Jiawen Wang$^{2}$,\\  Vivek Alumootil$^{3}$, Sangpil Kim$^{1,\dagger}$, and M. Khalid Jawed$^{2,\dagger}$%
\thanks{This research was funded in part by the US Department of Agriculture (Grant~\# 2024-67021-42528 and 2022-67022-37021), 
the Culture, Sports, and Tourism R\&D Program through the Korea Creative Content Agency grant funded by the Ministry of Culture, Sports and Tourism in 2024~(International Collaborative Research and Global Talent Development for the Development of Copyright Management and Protection Technologies for Generative AI, RS-2024-00345025),
and the Institute of Information \& Communications Technology Planning \& Evaluation (IITP) grant funded by the Korea government (MSIT) (No. RS-2019-II190079, Artificial Intelligence Graduate School Program, Korea University).}%
\thanks{$^{1}$ Department of Artificial Intelligence, Korea University, Seoul 02841, Republic of Korea. J. Jeong is also a visiting scholar at the University of California, Los Angeles, CA, USA. 
        {\tt\small \{jhwan, spk7\}@korea.ac.kr}}%
\thanks{$^{2}$ Department of Mechanical \& Aerospace Engineering, University of California, Los Angeles (UCLA), CA 90095, USA.
        {\tt\small \{tuananh.vu, radhalahoti, chiawenw\}@ucla.edu, khalidjm@seas.ucla.edu}}%
\thanks{$^{3}$ Department of Computer Science, University of California, Los Angeles (UCLA), CA 90095, USA.
        {\tt\small vivekalumootil@ucla.edu}}%
\thanks{$^{*}$ The authors contributed equally.}
\thanks{$^{\dagger}$ Co-corresponding authors.}
}

\begin{document}

\urlstyle{tt}

\maketitle

\begin{abstract}
Robotic pollination offers a promising alternative to manual labor and bumblebee-assisted methods in controlled agriculture, where wind-driven pollination is absent and regulatory restrictions limit the use of commercial pollinators. 
In this work, we present and validate a vision-guided robotic framework that uses data from an end-effector mounted RGB-D sensor and combines 3D plant reconstruction, targeted grasp planning, and physics-based vibration modeling to enable precise pollination.
First, the plant is reconstructed in 3D and registered to the robot coordinate frame to identify obstacle-free grasp poses along the main stem.
Second, a discrete elastic rod model predicts the relationship between actuation parameters and flower dynamics, guiding the selection of optimal pollination strategies.
Finally, a manipulator with soft grippers grasps the stem and applies controlled vibrations to induce pollen release.
End-to-end experiments demonstrate a 92.5\% main-stem grasping success rate, and simulation-guided optimization of vibration parameters further validates the feasibility of our approach, ensuring that the robot can safely and effectively perform pollination without damaging the flower.
To our knowledge, this is the first robotic system to jointly integrate vision-based grasping and vibration modeling for automated precision pollination.
\end{abstract}


\section{Introduction}
\label{sec:introduction}

Controlled environment agriculture (CEA), including greenhouses and indoor farms, is a sustainable solution to food production challenges worsened by climate change, labor shortages, and urbanization. While arable land growth is minimal (under 5\%), global food demand is projected to surge 70\% by 2050 \cite{FAO2050}. 
CEA offers efficient resource use (water, fertilizers, pesticides) and protection from environmental variability, but high operating costs—with labor comprising over 25-30\% of expenses for several crops~\cite{Ahamed2019}—limit its scalability, motivating the adoption of robotic systems to reduce this manual effort.

As production of major greenhouse crops like the tomato (\textit{Solanum lycopersicum})—the second most consumed fresh vegetable in the US—steadily increases, a key challenge in CEA is how to effectively pollinate in the absence of natural wind. The primary biological solution, bumblebees, is often unviable for two main reasons: their use is restricted or heavily regulated in states including California, Oregon, and Washington, and greenhouse lighting conditions can disorient them, reducing pollination efficiency~\cite{DeVries2020}. Consequently, growers rely on manual operation of mechanical vibration tools, such as vibrating wands and blowers, to induce flower vibration and facilitate pollen transfer. However, this manual approach is labor-intensive and costly, with expenses reaching 10,000–25,000 USD per hectare in Australia, highlighting the need for a more efficient, automated solution.

Robotic pollination has emerged as a promising alternative. Ground-based systems such as BrambleBee~\cite{Ohi2018} and commercial solutions by Arugga AI Farming ~\cite{Broussard2023} use air pulses or mechanical contact to pollinate flowers, achieving yields comparable to or exceeding manual methods. However, existing approaches are often proprietary, optimized for different crops, or risk damaging delicate flowers. Prior research also highlights challenges in navigating complex plant geometries, avoiding leaf obstructions, and ensuring precise, safe interaction with stems and flowers~\cite{Yuan2016}.

In this work, we present a novel robotic pollination framework designed to achieve efficient pollination while preventing flower damage. This framework integrates algorithm-based 3D plant skeletonization, collision-free grasp planning, and physics-based vibration analysis.
Our approach utilizes an end-effector mounted RGB-D sensor to perform generalizable 3D plant skeletonization and perception. This system secures 7-DoF safe grasp points and obstacle-free approach paths along the main stem, allowing the manipulator to safely grasp and induce vibration without damaging delicate flowers or thin stems.
Furthermore, we incorporate a Discrete Elastic Rod (DER) model to precisely analyze the relationship between stem actuation and flower motion. The results from this simulation guide the selection of optimal pollination parameters, which subsequently complements the experimental validation of the entire framework.

The main contributions of this work are as follows:
\begin{itemize}
    \item The first robotic system integrating vision-based grasp planning and physics-based vibration modeling for pollination, validated by a 92.5\% grasping success rate.
    \item Development of a novel 3D plant skeletonization technique enabling 7-DoF obstacle-free grasp selection for safe and generalized robotic manipulation.
    \item Utilization of a physics-based Discrete Elastic Rod model, experimentally validated to predict how flower dynamics vary with actuation parameters, thereby enabling a Sim-to-Real optimization framework for identifying optimal pollination strategies.
\end{itemize}

Section~\ref{sec:related} reviews prior work and highlights the limitations that motivate the framework architecture presented in Section~\ref{sec:method}. Sections~\ref{sec:skeletonization} and \ref{sec:rodmodel} detail the core methodology, introducing the 3D skeletonization algorithm for selecting 7-DoF obstacle-free grasp poses and the elastic rod model used for vibration dynamics analysis. Section~\ref{sec:experiments} presents the experimental setup and results for accuracy, grasp success, and vibration transfer. Section~\ref{sec:conclusion} summarizes the findings and discusses future work toward full greenhouse deployment.


\section{Related Work}
\label{sec:related}

\subsection{Robotic pollination in CEA.} 
Several robotic pollination systems have been developed for CEA. 
\textit{BrambleBee}~\cite{Ohi2018} employs a robotic arm with a soft brush to pollinate bramble flowers, integrating SLAM and visual servoing but lacking full autonomy as it depends on ArUco markers instead of real flowers.
\textit{StickBug}~\cite{smith2024design} improves efficiency by utilizing six independent manipulators; however, its pollination success rate remains low (49\%) due to frequent occlusions, as multiple arms obstruct the cameras. 
Both systems rely on direct physical contact between brushes and flowers, a limitation that risks permanent damage to fragile blossoms, such as those of tomatoes.
\textit{Arugga} AI~\cite{Arugga2022} has deployed autonomous robots in tomato greenhouses that use multiple air-jet nozzles for contactless pollination, avoiding flower damage and occlusion issues. However, as the system is proprietary and its algorithms and control strategies remain undisclosed, its reproducibility and broader research adoption are limited.

\subsection{Vision-based Plant Detection and Skeletonization}
Deep learning-based detection and segmentation models~\cite{10533619,ravi2024sam2} can, in theory, identify leaves, stems, and flowers in 2D images. However, due to frequent occlusions and overlapping structures inherent to plants, their accuracy is often low, and fine-grained organ-level details are not reliably captured. Moreover, these methods are limited to pixel-level segmentation and do not provide structural information, such as 3D geometry or articulation, which is critical for manipulation tasks.
Skeletonization has been used to extract structural information from plants, but most existing methods have been developed for humans~\cite{ren2024survey} or animals~\cite{li2025advances}. In agriculture, they have been applied mainly to trees and rigid or semi-rigid crops~\cite{cardenas2022modeling}, making them unsuitable for soft and flexible plants, such as tomato stems or peppers. Furthermore, studies on soft plants~\cite{wu2019accurate} are often trained on specific crops, limiting their generalizability.

\subsection{Plant Dynamics and Simulation}
Research on plant dynamics has often employed beam-based formulations within finite element frameworks to model a wide range of plant structures, from fir trees to wheat~\cite{chen2024analysis, moore2008simulating}. More recently, DER framework has been proposed as an efficient alternative for capturing plant dynamics~\cite{bergouDER}. However, early studies in this direction were limited to single-rod representations and did not account for the complex branching and joint elasticity characteristic of real plants. Recent extensions of the DER framework now address these limitations by modeling multiple interconnected elastic rods joined through compliant joints~\cite{choi2023dismech, MATdismech}, thereby enabling the realistic simulation of plant vibration dynamics relevant to pollination. Moreover, although a few studies have explored physics-based models for plant dynamics simulation~\cite{aghajanzadeh2022adaptive}, the crucial integration of such physically grounded models with robotic actuation and control planning remains largely unexplored, making the present work particularly valuable for bridging this gap.


\section{Robotic Pollination Methodology}
\label{sec:method}
\begin{figure*}
    \centering
    \includegraphics[width=1.00\linewidth]{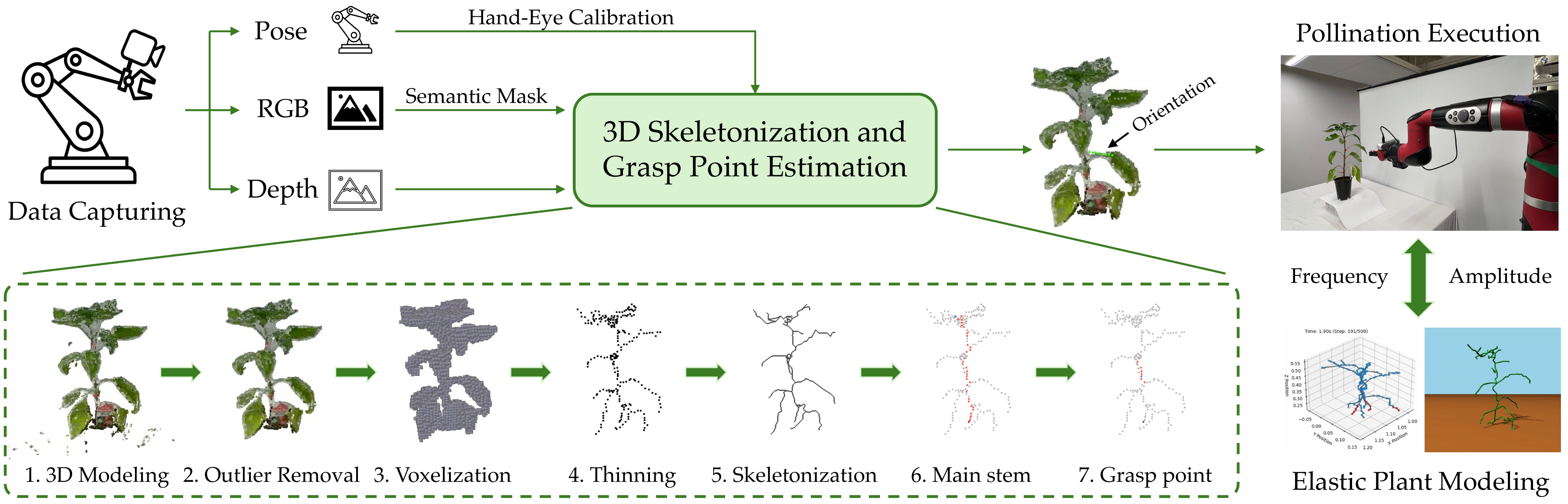}
    \caption{\textbf{Overview of the robotic pollination system.} The process begins by acquiring multi-view RGB-D images and corresponding 7-DoF end-effector poses to construct a 3D skeletonized plant model. From this model, an optimal grasp point and a collision-free trajectory are determined. A robotic arm then executes pollination by grasping and vibrating the stem. Elastic plant modeling simulations are utilized to examine the influence of grasp location and vibration parameters on flower motion, offering crucial insights to refine the strategy.}
    \label{fig:overall}
\vspace{-10pt}
\end{figure*}

This section details our two-stage method to determine where to grasp a plant stem and how to vibrate it for pollination (Figure~\ref{fig:overall}). First, a \textbf{(i) Vision-based pipeline} reconstructs a 3D plant skeleton from images, registers it to the robot's frame, and selects an obstacle-free 7-DoF grasp pose on the main stem. Second, an \textbf{(ii) Elastic rod-based plant dynamics model} uses this skeleton to model the plant’s dynamics, identifying the optimal grasp location and vibration parameters needed to maximize flower movement.


\subsection{3D Plant Skeletonization and Grasp Planning}
\label{sec:skeletonization}
We address the problem of modeling the 3D structure of a plant for robotic grasping. 
Our robotic system consists of a 7-DoF manipulator (Base $B$, Flange $F$, Gripper $G$), an RGB-D sensor ($C$), and the world frame ($W$).
In our setup, the base frame is aligned with the world frame ($W \equiv B$), and both the RGB-D camera and its extrinsic calibration are intrinsically and extrinsically calibrated.

\myparagraph{Robot-Camera Calibration.}
The robot's kinematic chain provides the forward transform $T_{B \leftarrow F} \in SE(3)$, describing the pose of the flange relative to the base. Through hand-eye calibration~\cite{88014}, the fixed transform $T_{F \leftarrow C}$ is ascertained. This calibration determines the camera's pose in the robot base frame for any given flange pose using the relationship:
\begin{align}
    T_{B \leftarrow C} = T_{B \leftarrow F} \cdot T_{F \leftarrow C}.
\end{align}
With the base and world frames aligned, this transform also defines the camera's world pose, denoted as $T_{W \leftarrow C}$.

\myparagraph{Semantic Depth Masking.}
To isolate the plant structure from background elements, including the pot and soil, semantic segmentation is performed on each RGB image. This process employs the \textit{Grounding DINO} zero-shot object detector~\cite{ren2024grounding} with the text prompt "leaves" to generate bounding boxes for the plant. These boxes are subsequently passed to the \textit{SAM2} model~\cite{ravi2024sam2segmentimages}, which produces a precise binary mask of the plant structure. By applying this mask to the corresponding depth map, all non-plant data is effectively filtered out, ensuring that only relevant structural points are used for 3D back-projection.

\myparagraph{Multi-View Fusion.}
We operate the manipulator to capture a dataset from multiple pre-defined viewpoints, where each capture consists of an RGB-D image and its corresponding 7-DoF pose. For each view $i$, the semantic masking procedure is first applied to yield a clean depth map containing only the plant. This masked depth is then back-projected into a partial point cloud in the camera's local frame. Each partial cloud is subsequently transformed into the world frame via its corresponding pose $T^{(i)}_{W \leftarrow C}$ and aggregated into a global model. To correct for minor drifts, this initial alignment is refined using the Iterative Closest Point (ICP) algorithm~\cite{zhang2021fast}, where the pose of each incoming cloud is updated as:
\begin{align}
    T'^{(i)}_{W \leftarrow C} = \Delta T^{(i)}_{\mathrm{ICP}} \, T^{(i)}_{W \leftarrow C},
\end{align}
with $\Delta T^{(i)}_{\mathrm{ICP}}$ being the corrective transform that minimizes the distance to the existing fused model, ensuring a globally consistent reconstruction.

\myparagraph{Point Cloud Preprocessing.}
The fused point cloud serves as input for the topological analysis. This data is preprocessed to ensure quality, beginning with a voxel grid downsampling to enforce uniform point density. The DBSCAN~\cite{10.5555/3001460.3001507} clustering algorithm is then applied to isolate the most significant connected component, effectively removing disconnected noise and reconstruction artifacts. This cleaned point cloud is finally converted into a high-resolution binary voxel grid that represents the volumetric occupancy of the plant.

\myparagraph{Topological Skeletonization.}
A one-voxel-thick topological skeleton is extracted from the binary voxel grid using a 3D thinning algorithm~\cite{lee1994building}. 
A graph, $G=(V, E)$, is constructed, where the vertices V are the skeleton points and the edges E are potential connections. These edges are proposed via a k-nearest neighbor (KNN) search and weighted by a custom cost function that prioritizes connections between thick, vertical segments, as identified by the grid's Euclidean Distance Transform (EDT). A Minimum Spanning Tree (MST) is computed on this weighted graph to yield a simplified, acyclic skeleton comprising only junctions and endpoints.

\myparagraph{Main Stem Grasp Pose Determination.}
The main stem is identified from the simplified skeleton in a two-stage process. First, a candidate stem is generated by finding the geodesic root-to-leaf path that maximizes the following score function, which prioritizes structural integrity and verticality:
\begin{align}
    S(\text{path}) = \sum_{e \in \text{path}} L_e \cdot \bar{r}_e^{\alpha} \left(1 + v_{\mathrm{bias}} \frac{|dz_e|}{L_e}\right),
\end{align}
where $L_e$, $\bar{r}_e$, and $dz_e$ are the length, mean radius, and vertical component of each edge $e$, respectively, while $\alpha$ and $v_{\mathrm{bias}}$ are pre-defined hyperparameters. Second, this candidate path is pruned to yield the final core stem by removing any segments that deviate from the primary vertical growth axis.
Once the core stem is identified, the optimal grasp point ($P_W$) is defined as the midpoint of its longest edge.
A collision-aware approach vector ($\vec{n}^*$) is chosen as the direction of least obstruction from nearby branches, ensuring a successful grasp by solving the following optimization:
\begin{align}
    \vec{n}^* = \underset{\vec{n} \perp \vec{v}_e, \|\vec{n}\|=1}{\arg\min} \left( \max_{j} |\vec{n} \cdot \vec{b}_j| \right).
\end{align}
Using $P_W$ as the target position, a 3D orientation is derived by aligning the gripper's approach axis with $\vec{n}^*$.
A rotation matrix is then formed by defining a default orientation for the gripper around this axis, which is converted to a quaternion to yield the final 7-DoF grasp pose for the robot flange.


\subsection{Elastic Rod-Based Plant Dynamics Model}
\label{sec:rodmodel}
\begin{figure}[t!]
\centering
\includegraphics[width=\columnwidth]{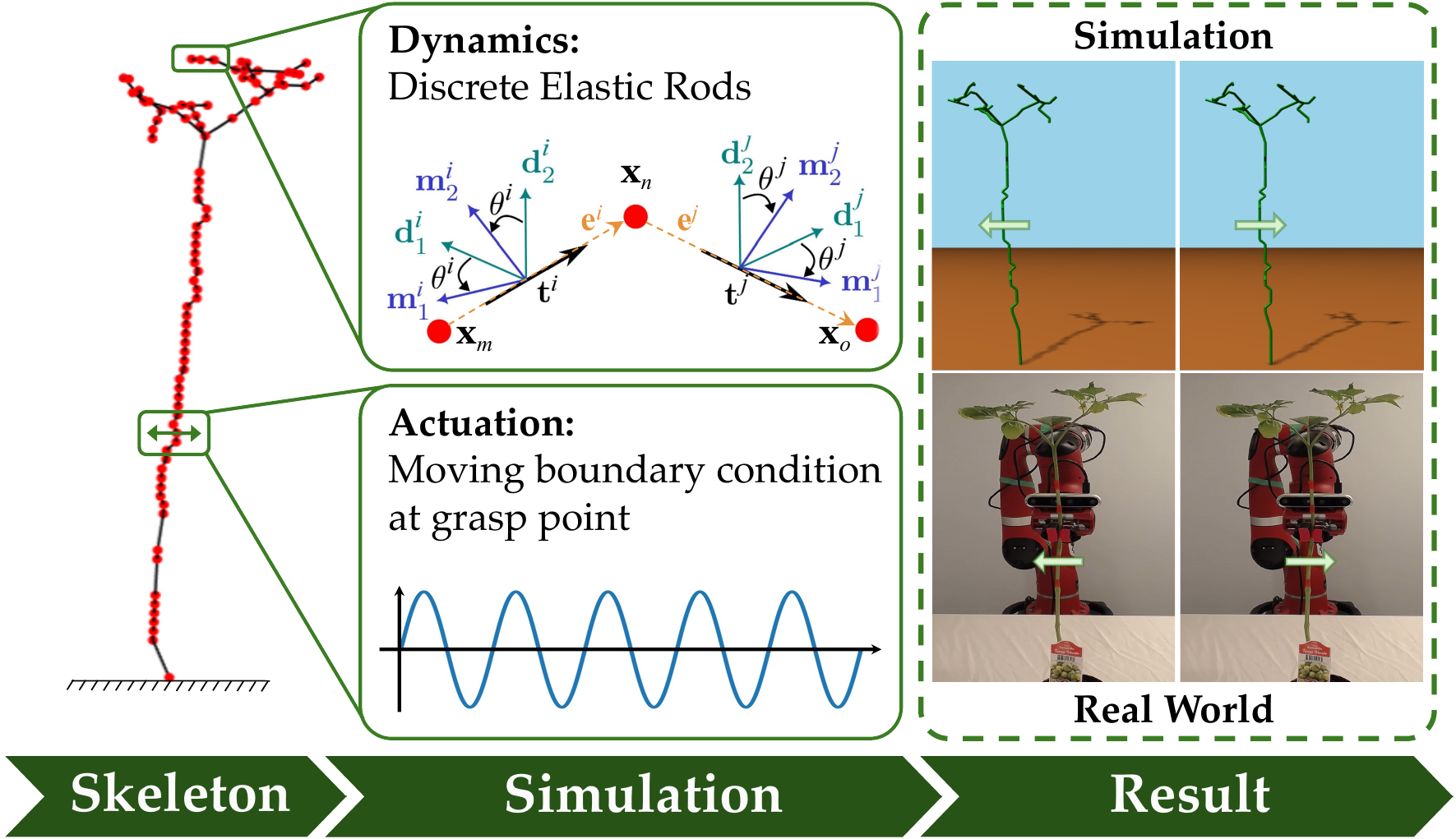}
\caption{\textbf{Simulation workflow using PyDiSMech~\cite{dismechPythonGithub}.} The plant is modeled as a network of slender rods (skeleton). A moving boundary condition is applied at the grasp point to mimic the vibration actuation, and the resulting dynamics are computed using the Discrete Elastic Rod method.}
\label{fig:sim}
\vspace{-10pt}
\end{figure}

To analyze the motion of the flowers during plant vibrations, we model the plant as a network of one-dimensional elastic rods. Each rod represents a slender plant segment, and the dynamics are simulated using the DER method~\cite{bergouDER}. While the DER framework was originally developed for individual rods, plant structures consist of branched, interconnected rods. To capture this network behavior, we employ the \textit{PyDiSMech} codebase~\cite{dismechPythonGithub}, which extends DER to simulate multiple connected rods while preserving elasticity at joint nodes, as described in MAT-DiSMech~\cite{MATdismech}. Figure~\ref{fig:sim}  illustrates the simulation workflow, including the discretized plant skeleton, the actuation applied at the grasp point, and the resulting plant dynamics obtained using PyDiSMech.

In DER, the structure is discretized into nodes and edges as shown in Figure~\ref{fig:sim}. The degrees of freedom typically include both the spatial positions of the nodes and the twist angles associated with edges, enabling three deformation modes: stretching, bending, and twisting. Elastic strain energy is stored in corresponding ``springs''~\cite{MATdismech}. At each time step, the governing \textit{equations of motion} are solved implicitly:
\begin{align}
\mathbf{M} \ddot{\mathbf{q}} = \mathbf{F}^{\text{elastic}} + \mathbf{F}^{\text{ext}},
\end{align}
where $\mathbf F^\textrm{elastic} = - \tfrac{\partial E^{\textrm{elastic}}}{\partial \mathbf q}$ is the elastic force derived from the elastic energy $E^{\textrm{elastic}}$, and $\mathbf{F}^{\text{ext}}$ contains external contributions such as gravity and damping.

In our plant dynamics model, we account only for stretching and bending deformations. Experiments indicate that twisting of stems and branches during vibration is negligible. Excluding twist, therefore, captures the essential plant behavior while reducing the system’s degrees of freedom and enhancing computational efficiency. Accordingly, the total elastic energy is
\begin{equation}\label{eq:E_elastic}
        E^{\textrm{elastic}} = \sum_{i=1}^S E^{\textrm{stretch}}_i + \sum_{i=1}^B E^{\textrm{bend}}_i,
\end{equation}
where $S$ and $B$ denote the numbers of stretching and bending springs, respectively. Each edge contributes a stretching spring, and each triplet of consecutive nodes (two adjacent edges) defines a bending spring, as shown in Figure~\ref{fig:sim}.

\paragraph{Stretching deformation} 
The axial strain of the $i$-th edge is
\begin{align}
\label{eq: stretch}
    \epsilon^{i} = \frac{\|\mathbf{e}^i\|}{\|\bar{\mathbf{e}}^i\|}-1,
\end{align}
where $\bar{\mathbf{e}}^i$ denotes the undeformed edge vector. The corresponding stretching energy is
\begin{align}
    E^{\textrm{stretch}}_i = \tfrac{1}{2} EA (\epsilon^{i})^2 \|\bar{\mathbf{e}}^{i}\|,
\end{align}
with stiffness $EA$.

\paragraph{Bending deformation} 
The curvature binormal at the central node of a bending spring is
\begin{align}
\label{equ:curvature}
    (\boldsymbol \kappa b)_k = \frac{2\,\mathbf{e}^{i}\times \mathbf{e}^{j}}{\|\mathbf{e}^{i}\|\|\mathbf{e}^{j}\|+\mathbf{e}^{i}\cdot \mathbf{e}^{j}},
\end{align}
from which scalar curvatures along the two material directors are extracted:
\begin{align}
\label{equ:curvature_material}
    \kappa^{(1)}_k &= \tfrac{1}{2}(\mathbf{m}^{i}_2+\mathbf{m}^{j}_2)\cdot (\boldsymbol \kappa b)_k, \\
    \kappa^{(2)}_k &= -\tfrac{1}{2}(\mathbf{m}^{i}_1+\mathbf{m}^{j}_1)\cdot (\boldsymbol \kappa b)_k,
\end{align}
where $\{\mathbf{m}^{i}_1, \mathbf{m}^{i}_2\}$ and $\{\mathbf{m}^{j}_1, \mathbf{m}^{j}_2\}$ denote the orthonormal material frames of edges $\mathbf{e}^i$ and $\mathbf{e}^j$ as shown in Figure~\ref{fig:sim}. The bending energy of the $k$-th spring is
\begin{align}
\label{equ:bending_energy}
    E^{\textrm{bend}}_k = \tfrac{1}{2}\,\frac{EI}{\Delta l_k} \left[(\kappa^{(1)}_k - \bar{\kappa}^{(1)}_k)^2 + (\kappa^{(2)}_k - \bar{\kappa}^{(2)}_k)^2 \right],
\end{align}
with $EI$ the bending stiffness, $\bar{\kappa}^{(1)}_k,\bar{\kappa}^{(2)}_k$ the natural (rest-state) curvatures, and $\Delta l_k = \tfrac{1}{2}(\|\bar{e}^{i}\| + \|\bar{e}^{j}\|)$ the Voronoi length. Full derivations of forces and Jacobians from these energy terms can be found in~\cite{jawed2018primer}.

The vibration was emulated by prescribing a time-varying displacement boundary condition at the grasp node. In some cases, to aid the convergence of the implicit time integration scheme, nearby nodes were also assigned small auxiliary ``guess'' displacements consistent with the expected oscillatory motion. However, these neighboring nodes remain unconstrained degrees of freedom, so the additional displacements merely serve as intelligent initialization values to facilitate numerical convergence and do not directly influence the simulated dynamics.

\myparagraph{Material Parameter Estimation.}
The material parameters—namely, density and Young’s modulus—were estimated using simple procedures. To calculate density, we cut relatively straight branches with uniform cross-sections, measured their mass, radius, and length, and computed the density from these values. An average density was obtained by repeating the procedure over multiple branches. Although significant variation was observed across samples, this approach provided a reasonable ballpark estimate.

To estimate Young’s modulus, we performed a vibration experiment by perturbing the stem and recording its natural oscillation frequency. Combining the measured frequency with the previously estimated density allowed us to infer the effective stiffness and, consequently, the Young’s modulus. In addition, we conducted a cantilever deflection experiment in which a known mass was attached to the free end of a horizontally held, relatively straight stem. The measured deflections were in order-of-magnitude agreement with predictions from Euler–Bernoulli beam theory, based on the estimated parameters.

We emphasize that these procedures provide only approximate, order-of-magnitude estimates of the material properties. The values may differ substantially from the actual parameters, as plant stems are fibrous, curved, non-homogeneous, and not strictly linear elastic.

\section{Experiments}
\label{sec:experiments}
This section presents experiments designed to validate our proposed robotic pollination framework. We begin by detailing the experimental setup and process used for real-world deployment. We then provide a qualitative evaluation of the 3D skeletonization performance, followed by a validation of our discrete elastic rod model against real-world plant dynamics. Finally, we demonstrate the effectiveness of the entire system through end-to-end trials, reporting a high grasp success rate on a physical robot.


\subsection{Experimental Setup}
\label{sec:experimentalsetup}
Our experimental platform consists of a 7-DoF Rethink Robotics Sawyer manipulator, equipped with an electric gripper attachment for the gentle handling of plants. For perception, an Intel RealSense D455 RGB-D camera is mounted on the robot's wrist in an eye-in-hand configuration, providing synchronized color images and depth maps. We evaluated this system on a set of eight pepper and two tomato plants. All computations are performed on an Intel Core i9-9900KF CPU, with the Grounding DINO~\cite{ren2024grounding} and SAM2 models~\cite{ravi2024sam2segmentimages} accelerated on an NVIDIA RTX 2080 Ti GPU.


\subsection{Experimental Protocol}
\begin{figure}[ht!]
\centering
\includegraphics[width=\columnwidth]{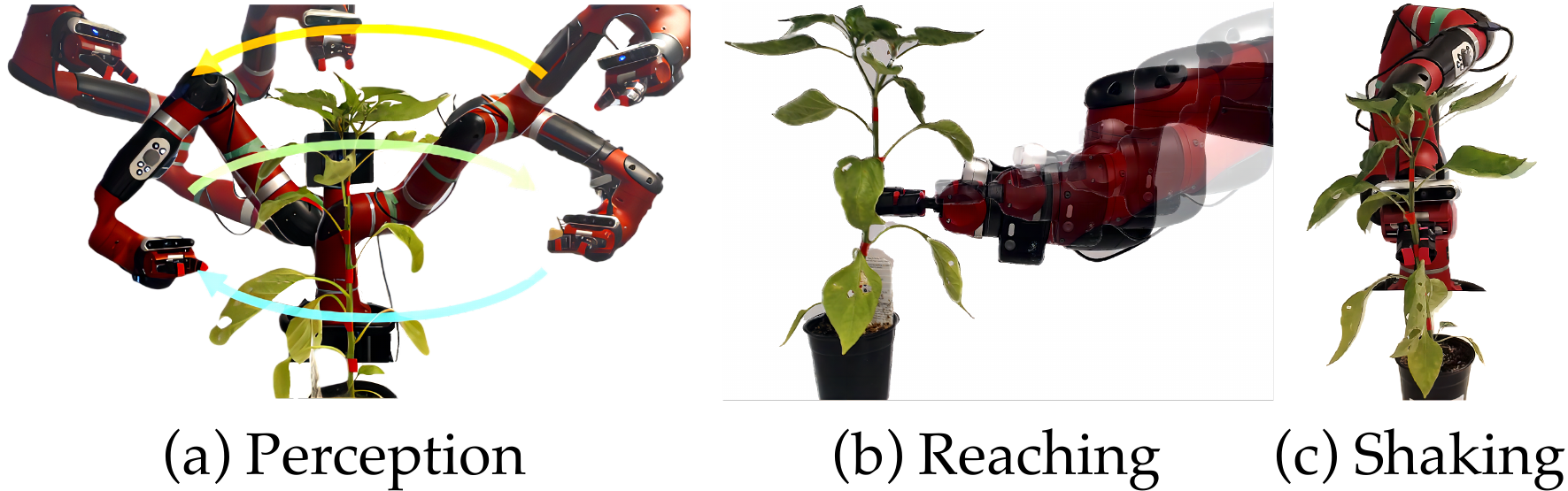}
\caption{The robotic pollination process, illustrated in three key stages: \textbf{Perception}, \textbf{Reaching}, and \textbf{Shaking}. In the Perception stage, the robot arm captures approximately 30 RGB-D images for 3D reconstruction. During the Reaching and Shaking stages, the arm reaches a target grasp position and shakes the plant's stem.}
\label{fig:robotprocess}
\end{figure}

Our experimental methodology involves a 90-second end-to-end trial per plant (Figure~\ref{fig:robotprocess}), followed by an offline policy refinement step. Each trial begins with (a) perception and planning, where the system captures ~30 images (approx. 45s) to compute a 3D skeleton and a 7-DoF grasp pose (approx. 30s). In the subsequent (b, c) robotic execution phase (approx. 15s), the manipulator grasps the stem and applies a shaking motion. The parameters for this motion are determined by an elastic rod model, which is optimized offline. Specifically, after each trial, the extracted skeleton is used to update the model and generate a refined actuation strategy for the next experiment.


\subsection{Results and Analysis}
\label{sec:evaluationMetrics}

\begin{figure}[ht!]
\vspace{-10pt}
\centering
\includegraphics[width=\columnwidth]{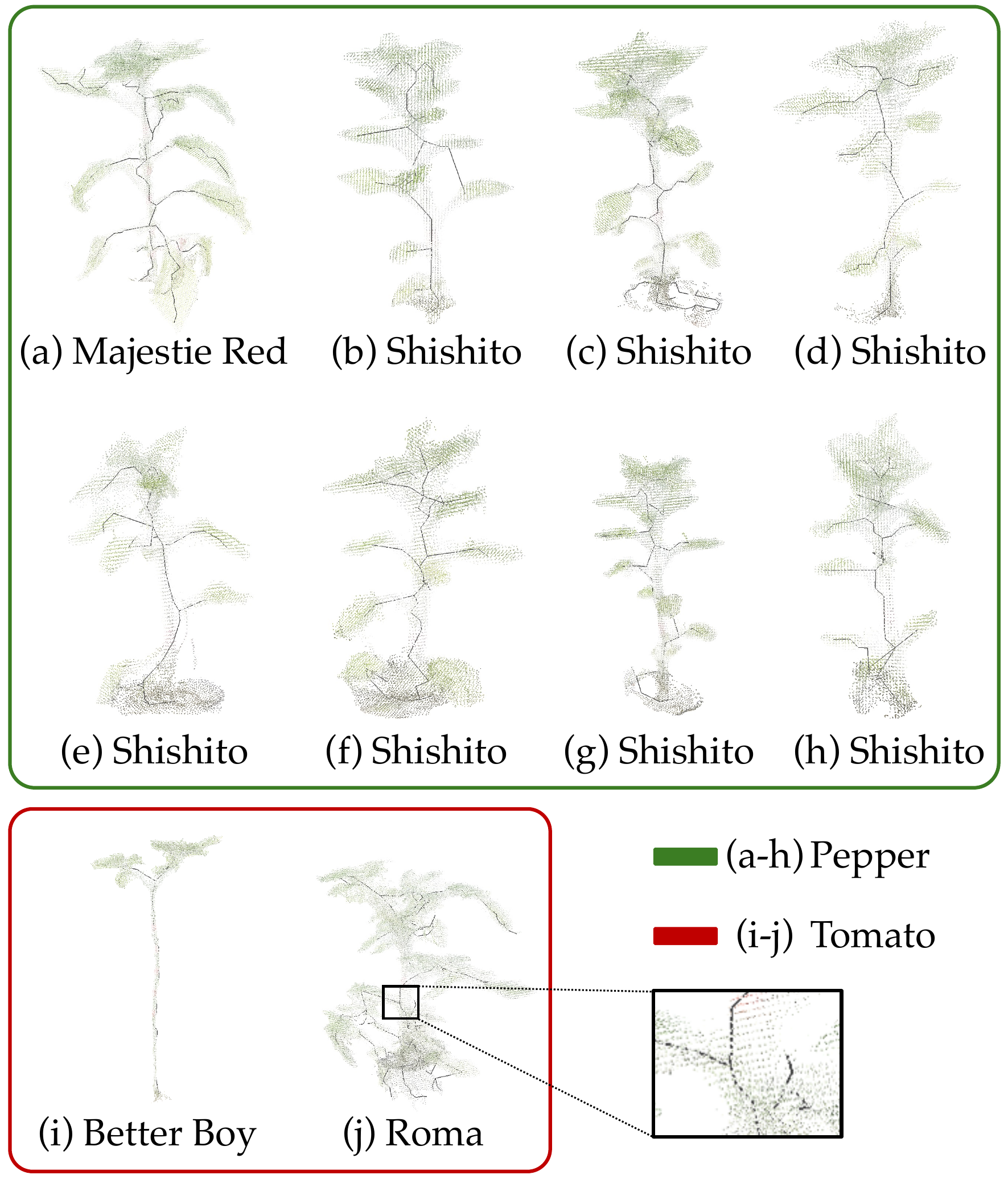}
\caption{Generalizability of our skeletonization algorithm, visualized by overlaying the skeletons on the 3D models of 10 diverse plants. The success across all plants using a single parameter set underscores the method's robustness.}
\label{fig:skeleton}
\end{figure}

\myparagraph{3D Skeletonization Accuracy.} 
Our skeletonization pipeline was evaluated on 10 plants with diverse morphologies, successfully generating skeletons that were well-aligned within the reconstructed 3D volume for all specimens, as shown in Figure~\ref{fig:skeleton}. 
Building on this success, the evaluation highlighted two areas for further improvement.
First, occasional segmentation bleed from \textit{SAM2}~\cite{ravi2024sam2segmentimages} into the planter pot was noted, which can interfere with root node identification. However, this is considered a minor issue in our target CEA environments where plant bases are obscured by soil or plastic mulch.
A second area for improvement relates to depth sensing.
Inaccurate depth data from the current sensor, which is unsuitable for near-field measurements, resulted in overestimated stem diameters and merged geometry, thereby reducing joint detection precision. The impact of input data quality was confirmed, as a dense point cloud generated from RGB images via \textit{COLMAP}~\cite{schoenberger2016sfm} yielded a more accurate skeleton. 
We therefore expect that upgrading to a depth sensor optimized for near-field applications, such as the \textit{Intel RealSense D405}(min distance: 7cm), would substantially enhance framework performance.

\myparagraph{Vibration Transfer Experiments.} 
To assess how mechanical vibrations applied at different points along the stem propagate to the flower, controlled vibration transfer experiments were conducted on two representative plants and compared against corresponding simulation results.

\begin{figure}[t!]
\centering
\includegraphics[width=\columnwidth]{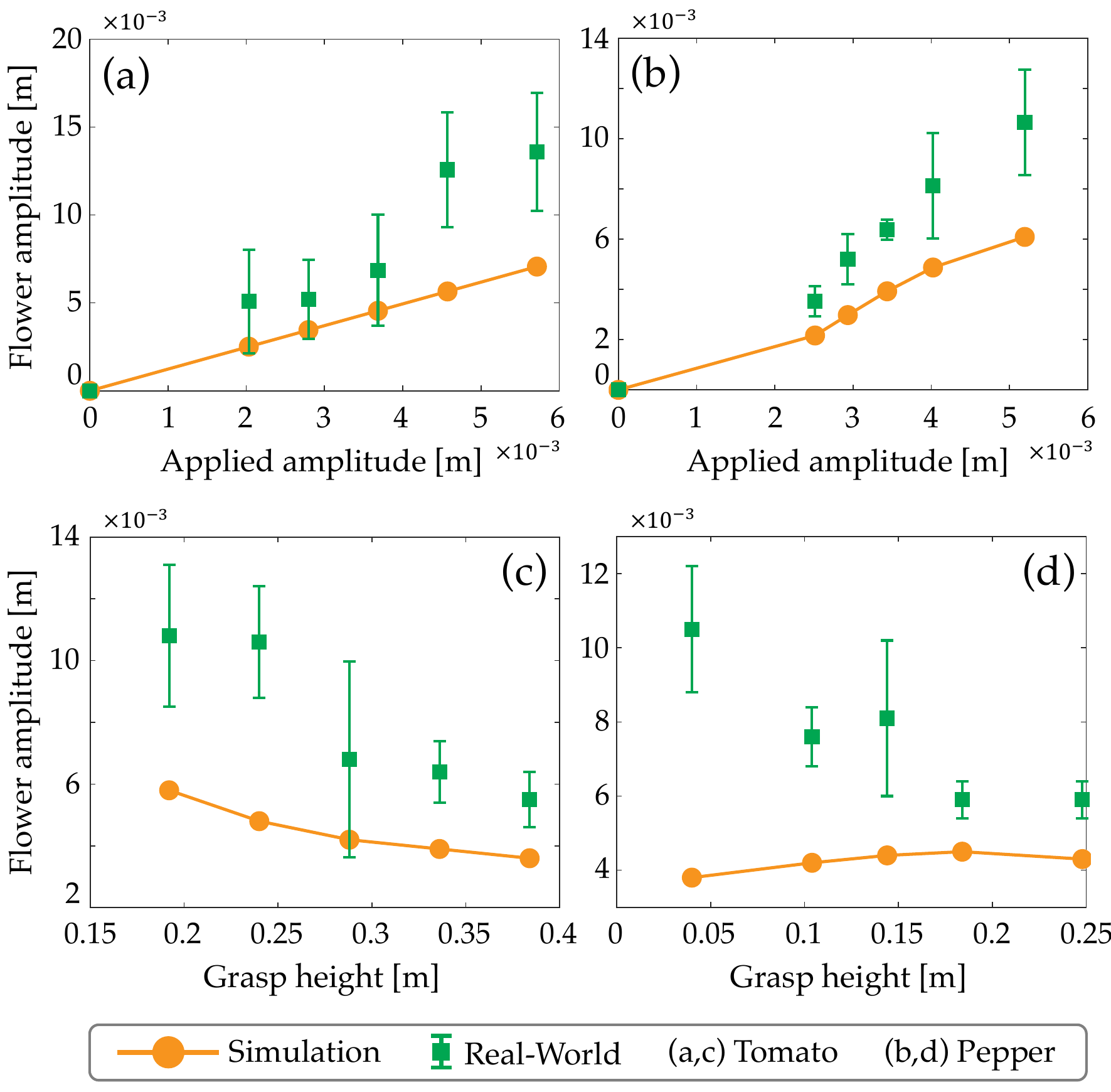}
\caption{
Comparison of experimental and simulated flower motion for Tomato (\textit{Better Boy}) and Pepper (\textit{Majestic Red}).
(a, b) Flower oscillation amplitude as a function of applied vibration amplitude, showing a clear positive correlation. 
(c, d) Flower oscillation amplitude as a function of grasping location (distance from root), showing that amplitude decreases as the grasp point moves closer to the flower.
}
\vspace{-10pt}
\label{fig:expt_vs_sim}
\end{figure}

Figures~\ref{fig:expt_vs_sim}(a, b) show the variation in flower oscillation amplitude with increasing vibration amplitude at the grasping location for Tomato (\textit{Better Boy}) and Pepper (\textit{Majestic Red}), respectively. 
Both experiments and simulations exhibit an approximately linear relationship between applied and resulting flower amplitudes, consistent with Euler--Bernoulli beam behavior. The simulations reproduce the observed trend with strong correlation ($r > 0.96$) but underpredict the amplitude magnitude by roughly 45\% on average, primarily because the model neglects stem tapering and local flexibility, which in reality reduce stiffness and amplify motion.

Figures~\ref{fig:expt_vs_sim}(c, d) show the dependence of flower amplitude on grasping location for the Tomato (\textit{Better Boy}) and Pepper (\textit{Majestic Red}) plants, respectively. In both cases, the experiments reveal that the flower amplitude decreases as the grasp point moves closer to the flower, reflecting the reduced effective vibrating length. The Tomato plant shows good quantitative agreement between experiment and simulation ($r \approx 0.92$, average error $\approx 50\%$), while the Pepper plant exhibits larger deviation and partial trend reversal ($r \approx -0.85$), likely due to its more complex branching structure, which was difficult to capture accurately through 3D reconstruction for simulation purposes.

\begin{figure}[!t]
\vspace{6pt}
\centering
\includegraphics[width=\columnwidth]{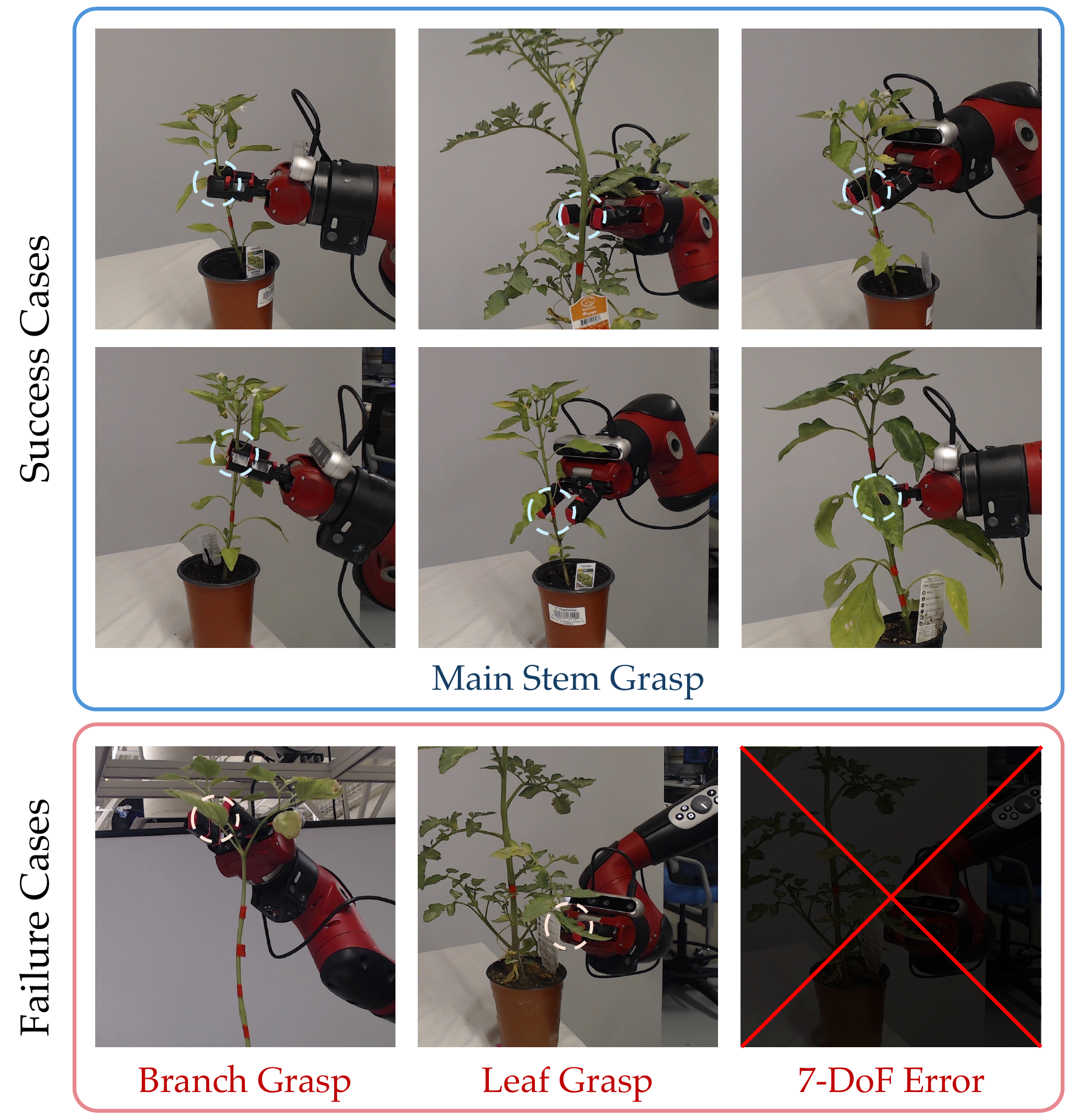}
\caption{Qualitative results of the real-world end-to-end experiments. The system achieved a 92.5\% success rate in grasping the main stem. The remaining 7.5\% of failures resulted from grasping a branch or leaf, or from the manipulator halting due to a 7-DoF pose error.}
\label{fig:cases}
\vspace{-5pt}
\end{figure}

\myparagraph{Grasping Performance in Real-World.} 
We conducted end-to-end experiments to evaluate the system's ability to successfully identify and grasp the main stem. To validate generalization, we performed 40 trials on 10 morphologically diverse plants, approaching each from four 90-degree viewpoints. A trial was deemed a success if the gripper correctly grasped the main stem, while a failure was recorded if it grasped a side branch or if the generated 7-DoF pose resulted in a manipulator error. 
The quantitative and qualitative results are summarized in Table~\ref{tab:realworld} and Figure~\ref{fig:cases}, respectively. Our robust framework approach demonstrated a high success rate of 92.5\%. This performance is attributed to the accurate 3D skeletonization, robust main stem identification, and the precise generation of 7-DoF grasp poses that are well-aligned with the robot's coordinate frame.

\begin{table}[!ht]
    \centering
    \begin{tabular}{ccc}
    \toprule
       \textbf{Plants}  & \textbf{Trials} & \textbf{Success Rate (\%)} \\
    \midrule
       Tomato  &  8  &  75.0\% \\
       Pepper  &  32   &  96.9\% \\
    \midrule
        Total  & 40  & 92.5\% \\
    \bottomrule
    \end{tabular}
    \caption{Success rate of robot grasping in real-world experiment. A high end-to-end success rate was achieved over 40 real-world grasping trials.}
    \label{tab:realworld}
\end{table}

\clearpage
\section{Conclusions}
\label{sec:conclusion}
We presented the first robotic system to jointly integrate vision-guided grasping with physics-based vibration modeling for autonomous pollination in CEA. Our framework leverages 3D skeletonization for precise grasp planning and a discrete elastic rod model to optimize vibration parameters, maximizing pollination efficiency. 
The simulations matched the experimental trends in flower motion with moderate amplitude discrepancies (40--55\%), suggesting that the model provides useful predictive insight for vibration-based pollination planning. 
Physical end-to-end experiments further validated the feasibility of this model-driven approach, achieving a 92.5\% main-stem grasping success rate and demonstrating that, guided by our dynamics model, the optimal vibration amplitude can be applied to successfully robotic pollination.
Future work will focus on scaling the system for greenhouse deployment, refining vibration parameters against fruit-set rates, and extending its adaptability to other crops. This model-guided strategy provides a clear pathway toward achieving highly efficient and reliable robotic pollination for sustainable food production.

\end{document}